\documentclass[lettersize,journal]{IEEEtran}
\usepackage{amsmath,amsfonts}
\usepackage{algorithmic}
\usepackage{algorithm}
\usepackage{array}
\usepackage[caption=false,font=normalsize,labelfont=sf,textfont=sf]{subfig}
\usepackage{textcomp}
\usepackage{stfloats}
\usepackage{url}
\usepackage{verbatim}
\usepackage{graphicx}
\usepackage{cite}
\usepackage{multirow}
\usepackage{hyperref}
\usepackage{booktabs}
\usepackage{colortbl}
\usepackage{hhline}

\begin{document}

\title{MC-PDD: Masked Corpus-Level Pretraining Data Detection for Black-Box Large Language Models}

        \author{Kaixin~Lan,~Mu~You,~Tao~Fang,~Binkai~Ou,~Lidia~S.~Chao,~Derek~F.~Wong
        \thanks{kaixin Lan, Lidia~S.~Chao and Derek F. Wong are with the Natural Language Processing \& Portuguese-Chinese Machine Translation (NLP$^2$CT) Laboratory, University of Macau, Macau SAR, China (e-mail: nlp2ct.kaixin@gmail.com; lidiasc@um.edu.mo; derekfw@um.edu.mo)}
        
         \thanks{Mu~You and Tao Fang are with the Institute of International Language Services Studies, Macau Millennium College, Macau SAR, China (e-mail: youmuafonso@mmc.edu.mo;  taofang@mmc.edu.mo)}

        \thanks{Binkai Ou is with the Innovation Research \& Development, BoardWare Information System Limited (e-mail: benson.ou@boardware.com)} 
        }



\maketitle

\begin{abstract}
Pretraining is fundamental to the development of Large Language Models (LLMs), yet the opacity of pretraining data complicates model analysis and raises ethical, legal, and fairness concerns. Detecting whether specific datasets were used during pretraining is, therefore, critical. Existing state-of-the-art methods typically rely on access to model probability distributions, making them unsuitable for closed-source LLMs that provide only input–output interfaces. To address this limitation, we introduce \textbf{M}asked \textbf{C}orpus-level \textbf{P}retraining \textbf{D}ata \textbf{D}etection (\textbf{MC-PDD}), a novel method inspired by the masked language modeling paradigm. MC-PDD masks highly specific tokens in each text and prompts the LLM to predict the missing content. It then assesses whether the difference in prediction hit rates between a candidate corpus and a reference non-member corpus is statistically significant. Based on this comparison, MC-PDD determines whether the candidate texts were likely included in the model’s pretraining data. Experimental results demonstrate clear and consistent differences in prediction hit rates between pretrained and unseen data across three datasets, for both open-source and closed-source LLMs. Despite operating under a stricter black-box setting, MC-PDD achieves performance comparable to existing detection methods. Our approach enables practical applications such as model auditing and data copyright verification using only standard API access. Upon acceptance, we will publicly release the code and datasets.
\end{abstract}


\section{Introduction}
\IEEEPARstart{T}{he} proliferation of large language models (LLMs) has revolutionized natural language processing \cite{NEURIPS2020_1457c0d6,fang2024llmclgec,jiao2023chatgpt,hendy2023good,pan2023preliminary,fang2023chatgpt,loem-etal-2023-exploring,fang-etal-2023-transgec}, yet it has also introduced critical challenges regarding data transparency. As pretraining datasets scale to trillions of tokens encompassing web-crawled corpora \cite{touvron2023llama,touvron2023llama2,openai2024gpt4}, the opacity of model development poses two significant challenges: (1) \textbf{data contamination} \cite{oren2023provingtestsetcontamination}, where benchmark samples inadvertently appear in pretraining data, leading to inflated performance estimates \cite{zhang2024pacostpairedconfidencesignificance, choi2025contaminatedbenchmarkquantifyingdataset}; and (2) \textbf{privacy and copyright risks}, as pretraining corpora may unintentionally include sensitive personal information or copyrighted materials \cite{carlini2021extractingtrainingdatalarge, cummings2024advancingdifferentialprivacyfuture}. These issues underscore the urgent need for reliable methods to detect whether specific datasets have been utilized during pretraining, a task that has become increasingly challenging with the prevalence of closed-source, API-based LLMs.

\begin{figure}[t]
    \centering
    \includegraphics[width=1.0\linewidth]{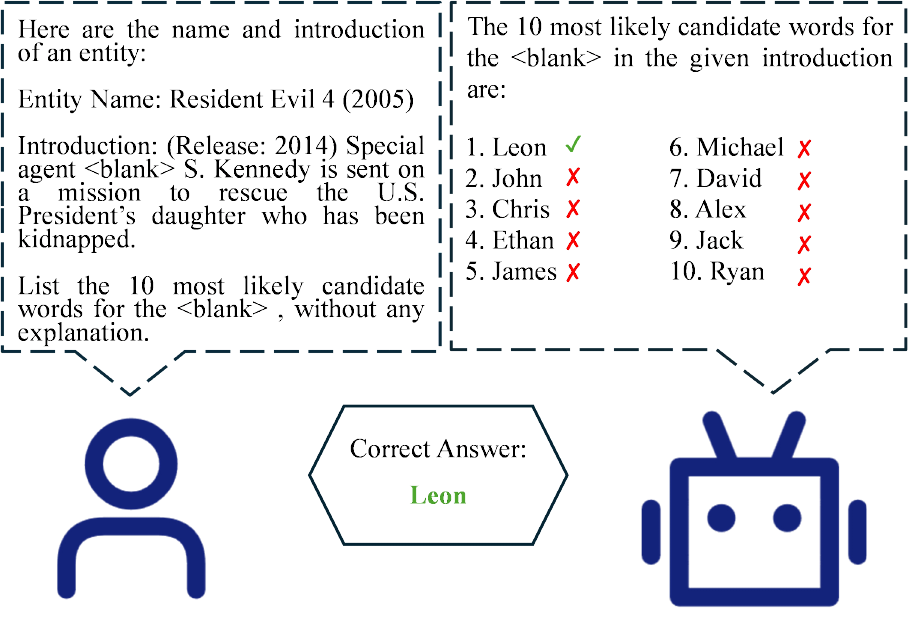}
    \caption{A typical model API call process in MC-PDD.}
    \label{fig:intro}
\end{figure}

Given the opacity of pretraining data in LLMs, determining whether specific datasets were used during pretraining is critical for model reliability, accountability, and regulatory compliance. However, current state-of-the-art (SOTA) detection methods, which primarily rely on white-box analyses of sample-level probability distributions \cite{carlini2021extractingtrainingdatalarge, shi2024detectingpretrainingdatalarge, zhang2025pretrainingdatadetectionlarge}, suffer from a fundamental limitation. Their dependence on access to model internals makes them incompatible with closed-source LLMs. As a result, a substantial gap remains between academic detection techniques and their practical applicability in real-world settings.

To address these issues, we propose \textbf{M}asked \textbf{C}orpus-level \textbf{P}retraining \textbf{D}ata \textbf{D}etection (\textbf{MC-PDD}), a novel black-box framework for pretraining data detection. Inspired by the masked language modeling paradigm, our approach leverages the observable differences between LLMs’ responses to pretraining data and unseen texts through masked token prediction analysis. Specifically, for each text, MC-PDD uses Term Frequency-Inverse Document Frequency (TF-IDF) \cite{sparck1972statistical} to identify the most specific token that is difficult to infer without prior knowledge and then prompts the model to predict it. Membership is inferred by testing whether the model’s average prediction hit rate is significantly higher on the candidate corpus than on a non-member reference corpus. The MC-PDD workflow is shown in Table~\ref{tab: MC-PDD}.

MC-PDD enables robust pretraining data detection using only input-output interactions. Its mechanism is illustrated in Figure~\ref{fig:intro}. When presented with a sentence containing a \texttt{<blank>} token, the model assigns a high probability to the token ``Leon'', the correct answer, suggesting that the example is likely drawn from the pretraining corpus where the model has gain memory of this passage. This result demonstrates MC-PDD’s ability to elicit memorized content from the model, even in complex and contextually rich settings.

We conduct a comprehensive set of experiments to evaluate the effectiveness and robustness of MC-PDD across both open-source and closed-source models. The results show a statistically significant gap in hit rates between member and non-member samples, even with as few as 100 samples. MC-PDD further demonstrates strong robustness on the BBC News and arXiv datasets, underscoring its practical applicability in real-world settings.

The main contributions are as follows:

(i) We propose MC-PDD, a novel black-box method for corpus-level pretraining data detection. We validate the effectiveness of MC-PDD through extensive experiments on three datasets and across both open-source and closed-source models, demonstrating that it achieves competitive performance compared to existing black-box methods, even under a stricter black-box setting that relies solely on input–output interactions.

(ii) We introduce two datasets, SteamMIA and arXivMIA-OLMo. SteamMIA consists of game metadata from Steam dated 2014 and 2024, while arXivMIA-OLMo comprises arXiv paper abstracts from February 2017 and February 2025, verified against the pretraining data of OLMo-7B-Instruct.

\begin{table}
\centering
\caption{Main flow of MC-PDD. MC-PDD masks the most specific token in each text, as determined by TF-IDF, and prompts the model to predict it. Membership is inferred by testing whether the model’s average prediction hit rate is significantly higher on the candidate corpus than on a non-member reference corpus.}\label{tab: MC-PDD}
\resizebox{.48\textwidth}{!}{
\begin{tabular}{ll} 
\toprule
\multicolumn{2}{l}{\textbf{MC-PDD}: Main Flow} \\ 
\midrule
\multicolumn{2}{l}{\textbf{Input}: Non-Member Corpus $C_n$, Candidate Corpus $C_c$, LLM $M$} \\
1 & \textbf{foreach} Corpus $C$ \textbf{in} $[C_n, C_c]$ \textbf{do~} \\
2 & ~~ Initialize counter $N$ = 0 \\
3 & \textbf{~~ foreach} Document $D$ \textbf{in} $C$ \textbf{do~ } \\
 & ~~~~ // Compute TF-IDF score and rank for each word in $D$ \\
4 & ~~~~ Mask word $w_n$= TF-IDF($D$, $[C_n, C_c]$).pop(0) \\
 & ~~~~ // Apply prompt template and get output \\
5 & ~~~~ Prompt $p$ = Prompt-Template($D$, $w_n$) \\
6 & ~~~~ Output $y$ = M($p$) \\
 & ~~~~ // Check output \\
7 & ~~~~ $N$ += 1 \textbf{if} $y$ contains $w_n$ \\
 & ~~ // Compute hit rate \\
8 & ~~ Hit rate $H$ = $N$ / len($C$) \\
9 & \textbf{if }$H_c$ is significantly higher than $H_n$:  \\
10 & ~~ $C_c$ is a member corpus \\
11 & \textbf{else:}  \\
12 & ~~ $C_c$ is a non-member corpus \\
\bottomrule
\end{tabular}
}

\end{table}

\section{Related Work}
\subsection{Pretraining Data Detection}
Pretraining data detection, a type of membership inference attack \cite{shokri2017membershipinferenceattacksmachine, zhang2025pretrainingdatadetectionlarge}, aims to determine whether a given test sample was included in a model’s pretraining data. Existing detection methods can be categorized based on model access: (i) \textbf{white-box} methods, which exploit internal model information such as weights, but are limited by restricted access to closed-source models; (ii) \textbf{black-box} methods, which rely on output token probabilities and are more practical in real-world scenarios.

Early approaches \cite{carlini2021extractingtrainingdatalarge} used perplexity for pretraining data detection, with specific techniques including Zlib and Lowercase. More recent advances have introduced increasingly effective techniques. \textbf{Min-K\% Prob} \cite{shi2024detectingpretrainingdatalarge} utilizes outlier token probabilities, while \textbf{Min-K\%++ Prob} \cite{zhang2025minkimprovedbaselinedetecting} enhances detection by normalizing token probabilities. \textbf{DPDLLM} \cite{zhou-etal-2024-dpdllm} leverages a reference model to extract features from generated and original text, achieving promising results. \textbf{DC-PDD} \cite{zhang2025pretrainingdatadetectionlarge} achieves SOTA performance by comparing token probability and frequency distributions. These methods focus on sample-level detection and often require access to token probabilities or reference models.
In contrast, our \textbf{MC-PDD} method targets corpus-level detection in a stricter black-box setting, using only input-output interfaces, and thereby further enhancing practicality. \cite{chang-etal-2023-speak} use a name-cloze task to assess model memorization, but the results vary widely across documents (accuracy ranging from 0.08 to 0.82). By operating at the corpus level, MC-PDD mitigates this instability and provides robust pretraining data detection.

\subsection{Membership Inference Attack}

Membership Inference Attacks (MIAs), introduced by \cite{shokri2017membershipinferenceattacksmachine}, aim to determine whether a given sample was included in a model’s training data through black-box queries.
Early research primarily focused on individual privacy \cite{carlini2021extractingtrainingdatalarge, cummings2024advancingdifferentialprivacyfuture} and data reconstruction \cite{gupta2022recoveringprivatetextfederated}, with applications mainly in computer vision \cite{10.1145/3460120.3484575, 10.1145/3448891.3448939, 10.1145/3372297.3417238, li2024blackboxmembershipinferenceattack}. These studies predominantly targeted fine-tuned models \cite{hisamoto-etal-2020-membership, jagannatha2021membershipinferenceattacksusceptibility, mattern-etal-2023-membership}. With the extension of MIAs to NLP, MIAs' heir applications broadened to include detecting pretraining data \cite{shi2024detectingpretrainingdatalarge, zhang2025pretrainingdatadetectionlarge}, probing for training data memorization \cite{nasr2023scalableextractiontrainingdata}, and data contamination \cite{oren2023provingtestsetcontamination}. Our work also targets pretraining data detection but operates under a more stringent black-box setting.

\subsection{Dataset Contamination}
Dataset contamination \cite{magar-schwartz-2022-data}, also known as benchmark leakage, occurs when the test dataset overlaps with the model’s pretraining corpus. This overlap can artificially inflate performance metrics and compromise the reliability of model evaluations \cite{brown2020languagemodelsfewshotlearners}, posing a significant challenge for conducting trustworthy assessments and developing AI-based benchmarks \cite{eriksson2025trustaibenchmarksinterdisciplinary}.
Research on dataset contamination detection is closely related to membership inference attacks; however, it primarily focuses on monitoring contamination in evaluation benchmarks \cite{oren2023provingtestsetcontamination, zhang2024pacostpairedconfidencesignificance, choi2025contaminatedbenchmarkquantifyingdataset}.

\section{Task Statement \& Challenge}
\subsection{Task Definition}
We formally define the task as follows: Given a black-box LLM with unknown pretraining data composition, the objective is to determine whether a candidate corpus was included in the model's pretraining data. 

\vspace{0.5em}
\noindent\textbf{Black-Box Setting.} Our approach fundamentally diverges from existing membership inference techniques by operating under a strictly surface-level interaction paradigm. Unlike approaches that leverage model internals (e.g., weight or token probabilities), we constrain the detector to only observe the input-output behavior of the LLM, mimicking the access level of standard API users.

\vspace{0.5em}
\noindent\textbf{Challenge 1: Acquisition of Test Datasets.}
While it is possible to construct test datasets based on model cutoff dates and data release timelines \cite{shi2024detectingpretrainingdatalarge}, a significant challenge arises from the lack of transparency regarding pretraining data. Since most model developers do not disclose details about their pretraining datasets, it becomes extremely difficult to ascertain whether data predating the model's cutoff time was genuinely included in the pretraining corpus. This creates a dilemma: access to the pretraining corpus is essential for constructing test datasets, yet such access is often unavailable.

\vspace{0.5em}
\noindent\textbf{Challenge 2: Detection Difficulties.}
\label{Detection Difficulties}
The detection process can be influenced by a model's language modeling capability and generalization ability. If a model has limited language modeling capacity, as is often the case with smaller-scale models, it may struggle to comprehend the detector's queries and to generate coherent responses, thereby hindering effective detection. However, models with strong generalization ability also introduce variability in their outputs, making them less predictable compared to direct database retrievals. This variability poses a significant challenge for detection, as distinguishing between responses derived from memorized pretraining data and those generated through generalization becomes difficult. Addressing these challenges requires crafting highly targeted queries that probe the pretraining data, as well as evaluation strategies designed to minimize the confounding effects of generalization.

\subsection{SteamMIA}
To address the challenges of constructing reliable test datasets under the black-box setting, we introduce the benchmark dataset, SteamMIA.
\begin{table}[t]
\centering
\caption{Distribution and release date of SteamMIA.}
\label{tab:steammia}
\begin{tabular}{lcc} 
\toprule
    & \multicolumn{1}{l}{Members} & \multicolumn{1}{l}{Non-Members}  \\ 
\hline
Size       & 468                         & 422                              \\
Release Date & 2014                        & 2024                             \\
\bottomrule
\end{tabular}
\end{table}

\vspace{0.5em}
\noindent\textbf{Dataset Construction.}
We adopt the model cut-off time approach proposed by \cite{shi2024detectingpretrainingdatalarge} to construct our dataset. Setting December 2023 as the temporal boundary, we collect publicly available game metadata from Steam. Games released in 2014 are categorized as \textit{members}, as their release predates that of most LLMs, while games released in 2024 are categorized as \textit{non-members}. The resulting dataset comprises 890 samples, with detailed statistics presented in Table~\ref{tab:steammia}. Each sample includes a game ID, name, and description obtained from the official Steam store webpage.

\vspace{0.5em}
\noindent\textbf{Why SteamMIA?}
In our experiments, we use the LLaMA-3.1-8B-Instruct model, as detailed in Section \ref{Model Selection}. Since its knowledge cutoff is December 2023, the previously proposed WikiMIA dataset \cite{shi2024detectingpretrainingdatalarge}, with a cutoff of January 2023, is not directly applicable. We verify this using SOTA detection methods. As shown in Table~\ref{tab:wikimia test}, these methods exhibit substantially degraded performance on WikiMIA, suggesting that the dataset’s non-member samples may have been included in LLaMA-3.1-8B-Instruct’s pretraining corpus. In contrast, SteamMIA yields results consistent with SOTA expectations, demonstrating its effectiveness for use with LLaMA-3.1-8B-Instruct.
\begin{table}[ht]
\centering
\caption{AUC score for WikiMIA and SteamMIA on LLaMA-3.1-8B-Instruct. The higher the better.}
\label{tab:wikimia test}
\begin{tabular}{lcc} 
\toprule
Method         & \multicolumn{1}{l}{WikiMIA} & \multicolumn{1}{l}{SteamMIA}  \\ 
\hline
PPL            & 0.5424                      & 0.7707                        \\
Lowercase      & 0.4772                      & 0.7602                        \\
Zlib           & 0.5660                      & 0.6644                        \\
Min-K\% Prob   & 0.5500                      & 0.7338                        \\
Min-K\%++ Prob & 0.5052                      & 0.5721                        \\
DC-PDD         & 0.5694                      & 0.7326                        \\
\bottomrule
\end{tabular}
\end{table}

\section{Methodology}
Table \ref{tab: MC-PDD} presents the main workflow of our methodology. Based on the analysis in section \ref{Detection Difficulties}, our method is designed as follows: first, we hypothesize that explicitly including an entity’s name and introductory information in the prompt can activate the model’s memory of that entity. Accordingly, our prompt template is structured as shown in Figure \ref{fig:entity_prompt_template}.

\begin{figure}[ht]
\centering
\small
\begin{tabular}{|p{0.45\textwidth}|}
\hline
\textbf{Prompt template} \\
\hline
Here are the name and introduction of an entity:\\
Entity Name: \{entity name\}\\
Introduction: \{introduction\}\\
List the 10 most likely candidate words for the \texttt{<blank>}, without any explanation:\\
\hline
\end{tabular}
\caption{Prompt template used for masked word prediction with entity name and introduction. Curly brackets \{\} denote placeholders to be completed.}
\label{fig:entity_prompt_template}
\end{figure}


The inclusion of ten candidate words in the model's response is designed to account for its generalization ability. Masked words are selected automatically using the TF-IDF method \cite{sparck1972statistical} (see pseudo code in Table \ref{tab: TF-IDF Algorithm}). 

The TF–IDF algorithm quantifies how distinctive a word is in a given document relative to a larger corpus. Its workflow consists of three main stages:

\vspace{0.5em}
\noindent\textbf{(i) Compute Term Frequency (TF).}
First, the algorithm processes a single input text ($W$). It counts the occurrences of each word and then normalizes these counts by dividing them by the total length of the text ($L_W$). This produces a dictionary ($D_{tf}$) representing how frequently each word appears in the document.

\vspace{0.5em}
\noindent\textbf{(ii) Compute Inverse Document Frequency (IDF).}
Next, the algorithm considers the entire corpus ($C$). It determines how many documents contain each word and, using the total number of documents ($L_C$), computes the IDF score as $\log(L_C / n)$, where $n$ is the number of documents containing the word. This step yields a dictionary ($D_{idf}$) that downweights common words and emphasizes rarer, more distinctive terms.
    
\vspace{0.5em}
\noindent\textbf{(iii) Compute Final TF-IDF Score.}
Finally, the algorithm combines the two measures. For each word in the TF dictionary, its TF value is multiplied by its corresponding IDF value. The result is a final dictionary ($D_{tf\text{-}idf}$) in which each word is assigned a weight reflecting its distinctiveness: higher scores indicate words that are frequent in the document but rare in the corpus.

In our approach, we compute the TF–IDF score for every word in the introduction and select the highest-scoring word as the candidate for masking. Choosing distinctive words is essential; if common words are selected, the task primarily evaluates the model’s general language modeling ability rather than its memorization of pretraining data. To assess the model’s response, we check whether the masked word appears among the ten generated candidates. If it does, the answer is marked correct; otherwise, it is labeled incorrect.

To ensure validity, member samples should be drawn from corpora that match the domain and stylistic characteristics of the test samples, minimizing irrelevant variables. While using the model's original pretraining data would be ideal, practical constraints often make this infeasible. As an alternative, we propose sourcing publicly available web-based texts, preferably from the same origin but dated earlier (e.g., by several years), as these are likely to have been incorporated into the model's pretraining corpus.
After evaluating the entire dataset, we compute the average hit rate and compare the results between member and test samples. A statistically significant difference between the two indicates that the test samples are unlikely to have been part of the model’s pretraining data.


\begin{table}
\centering
\caption{Pseudo code of TF-IDF Algorithm.}
\label{tab: TF-IDF Algorithm}
\begin{tabular}{ll} 
\toprule
\multicolumn{2}{l}{\textbf{TF-IDF}: Compute TF} \\ 
\midrule
\multicolumn{2}{l}{\textbf{Input}: Input Text $W$} \\
\multicolumn{2}{l}{\textbf{Result}: Dictionary of TF $D_{tf}$} \\
1 & Initialize $D_{tf}$ as empty dictionary \\
2 & Get length of Text as $L_W$ \\
3 & \textbf{foreach} word $w$ \textbf{in} $W$ \textbf{do~ } \\
 & ~~ $D_{tf}$[$w$] += 1 \\
4 & \textbf{foreach} word $w$ in $W$ \textbf{do} \\
 & ~~ $D_{tf}$[$w$] /= $L_W$ \\
5 & \textbf{return} $D_{tf}$ \\
 &  \\ 
\hline
\multicolumn{2}{l}{\textbf{TF-IDF}: Compute IDF} \\ 
\hline
\multicolumn{2}{l}{\textbf{Input}: Input Corpus $C$ } \\
\multicolumn{2}{l}{\textbf{Result}: Dictionary of IDF $D_{idf}$} \\
1 & Initialize $D_{idf}$ as empty dictionary \\
2 & Initialize $D_{word}$ as empty dictionary \\
3 & Get length of Corpus $C$ as $L_C$ \\
4 & \textbf{foreach} document $c$ \textbf{in} $C$ \textbf{do~ } \\
5 & ~~\textbf{foreach} word $w$ \textbf{in} $c$ \textbf{do~} \\
 & ~~~~ $D_{word}$[$w$] += 1 \\
6 & \textbf{foreach} word $w$, count $n$ \textbf{in} $c$ \textbf{do~} \\
 & ~~$D_{idf}$[$w$] = log($L_C$ / $n$) \\
7 & \textbf{return} $D_{idf}$ \\
 &  \\ 
\hline
\multicolumn{2}{l}{\textbf{TF-IDF}: Compute TF-IDF} \\ 
\hline
\multicolumn{2}{l}{\textbf{Input}: Dictionary of TF~$D_{tf}$, and IDF $D_{idf}$} \\
\multicolumn{2}{l}{\textbf{Result}: Dictionary of TF-IDF $D_{tf-idf}$} \\
1 & Initialize $D_{tf-idf}$ as empty dictionary \\
2 & \textbf{foreach} word $w$ \textbf{in} $D_{tf}$ \textbf{do~} \\
 & ~~$D_{tf-idf}$[$w$] = $D_{tf}$[$w$] * $D_{idf}$[$w$] \\
3 & \textbf{return} $D_{tf-idf}$ \\
\bottomrule
\end{tabular}
\end{table}

\section{Experimental Settings}

\vspace{0.5em}
\noindent\textbf{Model Selection.}
\label{Model Selection}
As previously mentioned, it is crucial to assess whether the model possesses sufficient language capability to comprehend queries and generate coherent responses. In our preliminary experiments, we examine 7B-scale models used in prior works, such as Pythia \cite{biderman2023pythiasuiteanalyzinglarge}, GPT-NeoX \cite{black2022gptneox20bopensourceautoregressivelanguage}, and LLaMA \cite{touvron2023llamaopenefficientfoundation}. However, as these models are base versions without fine-tuning, they lack instruction-following abilities and struggle to provide consistent answers. To address this, we test their instruction-tuned variants (where available) and select LLaMA-3.1-8B-Instruct \cite{grattafiori2024llama3herdmodels}, which demonstrates the most stable performance for our testing purposes. 

To further investigate the effectiveness of our approach on closed-source models, we evaluate several models via their official APIs, 
including Claude-3-Haiku\footnote{\href{https://www.google.com/url?sa=t&source=web&rct=j&opi=89978449&url=https://www-cdn.anthropic.com/de8ba9b01c9ab7cbabf5c33b80b7bbc618857627/Model_Card_Claude_3.pdf&ved=2ahUKEwjIpKaPmMWLAxWdjq8BHfv9K_cQFnoECAkQAQ&usg=AOvVaw3IPcOaw3w4_mjNiyMCY4x6}{Claude-3-Haiku}}, DeepSeek-V3-Chat \cite{deepseekai2024deepseekv3technicalreport}, Gemini-1.5-Flash \cite{geminiteam2024gemini15unlockingmultimodal}, GPT-4o-Mini\footnote{\href{https://openai.com/index/gpt-4o-mini-advancing-cost-efficient-intelligence/}{GPT-4o-Mini}}, and LLaMA-3-70B \cite{grattafiori2024llama3herdmodels}.

We also test our method using the open-source OLMo-7B-Instruct model \cite{groeneveld2024olmoacceleratingsciencelanguage} on a dataset that includes arXiv articles, some of which were part of the model's pretraining data.
The knowledge cutoff dates for all models (if available) are presented in Table \ref{tab:cutoff}, with information sourced from three websites\footnote{\url{https://huggingface.co/models}}\footnote{\url{https://github.com/HaoooWang/llm-knowledge-cutoff-dates}}\footnote{\url{https://agicto.com/llm-leaderboard}}.

\vspace{0.5em}
\noindent\textbf{Additional Datasets.}
To mitigate potential biases stemming from the unique characteristics of SteamMIA, we incorporated two additional datasets: the widely used BBC\_News\_Alltime \cite{li2024latesteval} and ArXiv\_Alltime \cite{li2024latesteval}. These datasets comprise BBC news articles and arXiv papers\footnote{\url{https://arxiv.org/}}, respectively, spanning from 2017 to the present and updated monthly. Using these two datasets, we conducted validation experiments with Qwen2.5-14B-Instruct and OLMo-7B-Instruct to assess the robustness of MC-PDD. As OLMo-7B-Instruct is fully open-source, its publicly documented training data indicate that ArXiv papers were included during pretraining.

\vspace{0.5em}
\noindent\textbf{Implementation details.}
For experiments involving the open-source models LLaMA-3.1-8B-Instruct, Qwen2.5-14B-Instruct, and OLMo-7B-Instruct, we use an NVIDIA A40 GPU (48 GB). The output parameters are set as follows: a maximum length of 200 tokens and a temperature of 0.1, with all other parameters kept at their default values. In the ablation experiments, we randomly select 200 non-member samples and perform five epochs of LoRA training on LLaMA-3.1-8B-Instruct. The training is conducted using code from LLaMA-Factory\footnote{\href{https://github.com/hiyouga/LLaMA-Factory/tree/main}{LLaMA-Factory}}, with the following configuration: \texttt{num\_train\_epochs = 5.0}, \texttt{save\_strategy = epoch}, \texttt{learning\_rate = 1.0e-4}, \texttt{bf16 = true},  and all other parameters set to their default values.

\begin{table}[t]
\centering
\caption{Models’ knowledge cutoff dates. An asterisk (*) denotes a knowledge cutoff later than December 2023.}
\label{tab:cutoff}
\resizebox{.45\textwidth}{!}{
\begin{tabular}{lc}
\hline
\toprule
Model Name              &Knowledge Cutoff \\ 
\cmidrule(lr){1-2}
Gemini-1.5-Flash\textsuperscript{*}         & 2024.05                              \\ 
LLaMA-3.1-8B-Instruct   & 2023.12                              \\ 
LLaMA-3-70B             & 2023.12                              \\ 
GPT-4o-Mini             & 2023.10                              \\ 
Claude-3-Haiku-20240307 & 2023.08                              \\ 
Deepseek-V3-Chat        & 2023.05                              \\
OLMo-7B-Instruct        & 2023.03                              \\
\bottomrule
\end{tabular}}
\end{table}

\begin{table*}[!t]
\centering
\caption{Hit rate and confidence interval upper bound on SteamMIA across the six evaluated models. All models with a knowledge cutoff date prior to December 2023 exhibit significant differences in hit rate. Data highlighted in red indicate non-significant hit rate difference. An asterisk (*) denotes models with a knowledge cutoff date later than December 2023. In this paper, ``CI Upper Bound'' refers to the upper limit of 95\% confidence interval. Except for Gemini-1.5-Flash, the hit rate difference remains significant for all models, with the smallest difference exceeding 9\%.}
\label{tab:main result}
\resizebox{0.95\textwidth}{!}{
\begin{tabular}{ccclc} 
\toprule
LLM   & \multicolumn{1}{l}{Hit Rate on Non-members} & Hit Rate on Members & Hit Rate Difference  & \multicolumn{1}{l}{CI Upper Bound}  \\ 
\hline
\multicolumn{1}{l}{LLaMA-3.1-8B-Instruct} & 36.26\%                                     & 48.50\%             & -12.24\% ($P$=0.0001) & -7\%                                        \\
Claude-3-Haiku                            & 40.54\%                                     & 54.50\%             & -13.96\% ($P$=0.0001)  & -8\%                                        \\
Deepseek-V3-Chat                          & 49.77\%                                     & 61.85\%             & -12.08\% ($P$=0.0005)  & -7\%                                        \\
GPT-4o-Mini                               & 37.16\%                                     & 46.21\%             & -9.05\% ($P$=0.0033)   & -4\%                                        \\
LLaMA-3-70B                               & 43.92\%                                     & 58.06\%             & -14.14\% ($P$=0.0001)  & -9\%                                        \\
Gemini-1.5-Flash\textsuperscript{*}       & 44.37\%                                     & 46.32\%             & {\cellcolor[rgb]{0.827,0.725,0.718}}-1.95\% ($P$=0.2808)   & {\cellcolor[rgb]{0.827,0.725,0.718}}4\%                                         \\
\bottomrule
\end{tabular}}
\end{table*}

\begin{table}
\centering
\caption{Accuracy and AUC scores on SteamMIA with LLaMA-3.1-8B-Instruct. The highest scores are shown in bold.}
\label{tab:main result compare}
\resizebox{.75\linewidth}{!}{
\begin{tabular}{lcc} 
\toprule
Method & Accuracy & AUC \\ 
\hline
PPL & 0.7113 & \textbf{0.7707} \\
Lowercase & 0.7038 & 0.7602 \\
Small Ref & \textbf{0.7794} & 0.7304 \\
Zlib & 0.6360 & 0.6644 \\
Min-K\% Prob & 0.6782 & 0.7338 \\
Min-K\%++ Prob & 0.5683 & 0.5721 \\
DC-PDD & 0.6877 & 0.7326 \\
MC-PDD & 0.7223 & 0.7595 \\
\bottomrule
\end{tabular}
}
\end{table}

\section{Results}
\subsection{Result on SteamMIA}
Table \ref{tab:main result} presents the performance of MC-PDD on SteamMIA across the six evaluated models. A key observation is that, except for Gemini-1.5-Flash, all models exhibit a minimum hit rate difference of 9\% between member and non-member samples. This difference is statistically significant, demonstrating the effectiveness of our method. In contrast, Gemini-1.5-Flash shows a hit rate difference of less than 2\%, which we attribute to its knowledge cutoff in May 2024, suggesting potential contamination of the non-member samples.
Additionally, LLaMA-3-70B, which shares the same architecture as LLaMA-3.1-8B-Instruct, achieves a higher hit rate. This indicates that model scaling (or improved language modeling capability) enhances the ability to predict masked tokens. Notably, DeepSeek-V3-Chat attains the highest hit rate across both member and non-member samples. This is likely due to its massive 671B parameters. Furthermore, the significant hit rate difference observed in DeepSeek-V3-Chat demonstrates that MC-PDD is also effective for models with a Mixture of Experts (MoE) architecture.

Subsequently, we compare MC-PDD with existing methods. We randomly select 122 samples from the SteamMIA-2014 dataset as the reference set. The remaining data are divided into two positive groups (seen) and three negative groups (unseen), each containing 150 samples. Each group is then tested on MC-PDD against the reference set using LLaMA-3.1-8B-Instruct. The experiment is repeated ten times, and the average accuracy and AUC are reported.
As shown in Table \ref{tab:main result compare}, MC-PDD achieves comparable performance to the best existing methods without requiring any access to the model’s internal weights or prediction probabilities. This confirms the practical applicability of our proposed approach.

\subsection{Result on BBC News}
We sampled news articles published in March of each year from BBC\_News\_Alltime. These samples were evaluated on MC-PDD using the Qwen2.5-14B-Instruct model for experimental validation. Statistical significance analyses were performed to compare the March 2025 results with those from earlier periods, with the outcomes summarized in Table \ref{tab: Qwen2.5-bbc-time}.

Although explicit documentation regarding the model's knowledge cutoff date is not publicly available, we can assert that the March 2025 data represent non-member samples, as Qwen2.5-14B-Instruct was officially released in July 2024.
Our analysis reveals a consistent and statistically significant hit rate difference between March 2025 and more distant years (2017–2020 and 2022), demonstrating the robustness of our method. In contrast, no significant differences were observed for 2023 and 2024.

We propose two potential explanations. First, these recent data may not have been included in the model's pretraining corpus, as dataset compilation typically precedes model release by a substantial margin. Second, the model's memorization patterns may vary across different training phases, suggesting that knowledge retention could be time-dependent. Further investigation into the model's training methodology is needed to validate this hypothesis and clarify the underlying mechanisms.

\begin{table}[ht]
\centering
\caption{MC-PDD results on BBC News using Qwen2.5-14B-Instruct. Comparisons are made between March 2025 data and data from March 2024 and earlier.}
\label{tab: Qwen2.5-bbc-time}
\begin{tabular}{cccc} 
\toprule
Date & Hit rate & CI Upper Bound & $P_{value}$ \\ 
\hline
2017-03 & 36.75\% & -0.03 & 0.0005 \\
2018-03 & 36.31\% & -0.04 & 0.0002 \\
2019-03 & 36.16\% & -0.03 & 0.0001 \\
2020-03 & 35.78\% & -0.03 & 0.0003 \\
2021-03 & 31.72\% & 0.01 & 0.1497 \\
2022-03 & 34.51\% & -0.02 & 0.0015 \\
2023-03 & 31.51\% & 0.01 & 0.1622 \\
2024-03 & 29.03\% & 0.03 & 0.8090 \\
2025-03 & 30.08\% & - & - \\
\bottomrule
\end{tabular}
\end{table}

To validate our word selection strategy, we conducted controlled experiments using BBC News data from February 2017 as member samples and February 2025 as non-member samples. As shown in Table \ref{tab: Qwen2.5-random}, we compared two masking approaches: ``Max'' (masking the word with the highest TF-IDF score) and ``Random'' (masking a word at random). The results reveal a clear distinction between the two methods: (i) the ``Max'' selection demonstrated statistically significant differentiation between member and non-member samples, validating our strategy; (ii) in contrast, the ``Random'' selection yielded high hit rates (approaching 60\%) while showing no significant differentiation, effectively reducing the task to a generic language modeling test. This contrast empirically confirms that random word selection fails to discriminate membership status, whereas our TF-IDF-based approach effectively captures distinctive patterns reflecting the model’s memorization of its pretraining data.

\begin{table}[ht]
\centering
\caption{MC-PDD results on BBC News with different word selection strategies using Qwen2.5-14B-Instruct.}
\label{tab: Qwen2.5-random}
\begin{tabular}{ccccc} 
\toprule
Strategy & Date & Hit rate & CI Upper Bound & $P_{value}$ \\ 
\hline
\multirow{2}{*}{Max} & 2017-02 & 40.78\% & \multirow{2}{*}{-0.08} & \multirow{2}{*}{0.0001 } \\
 & 2025-02 & 30.60\% &  &  \\
\multirow{2}{*}{Random} & 2017-02 & 59.77\% & \multirow{2}{*}{0.02} & \multirow{2}{*}{0.3814} \\
 & 2025-02 & 59.31\% &  &  \\
\bottomrule
\end{tabular}
\end{table}

\subsection{Result on ArXiv Papers}
To further validate MC-PDD, we applied it to the ArXiv\_Alltime dataset using OlMo-7B-Instruct. We selected data from February 2017, February 2018, and February 2025, and evaluated multiple word selection strategies and selection sizes, following prior experimental settings. Since academic papers often contain a large number of mathematical symbols, formulas, and figures that hinder accurate model responses, we used regular expressions to select words composed solely of English letters.

As shown in Table \ref{tab: Olmo-Arxiv}, under the ``Max'' strategy, the hit rates for member and non-member samples differ significantly, whereas this difference becomes insignificant when masked words are selected at random. This demonstrates the effectiveness of our method in real-world scenarios. Notably, under the ``Max'' strategy, the upper bound of the 95\% confidence interval is close to zero, suggesting that detecting this category of texts (academic papers) approaches the robustness limit of MC-PDD.


\begin{table*}[!ht]
\centering
\caption{MC-PDD results on ArXiv papers using different word selection sizes and selection strategies with OlMo-7B-Instruct. Non-members correspond to data from February 2025. Top-$n$ denotes the number of specificity words selected per paper.}
\label{tab: Olmo-Arxiv}
\resizebox{0.95\textwidth}{!}{
\begin{tabular}{ccccccc} 
\toprule
Strategy & Date & Top-n & Hit Rate on Members & Hit Rate on Non-members & CI Upper Bound & $P_{value}$ \\ 
\hline
\multirow{6}{*}{Max     } & \multirow{3}{*}{2017-02} & 1 & 6.67\% & 3.05\% & -0.01 & 0.0030 \\
 &  & 3 & 7.05\% & 4.70\% & -0.01 & 0.0023 \\
 &  & 5 & 6.83\% & 4.94\% & -0.01 & 0.0013 \\ 
\cmidrule(lr){2-7}
 & \multirow{3}{*}{2018-02} & 1 & 5.14\% & 3.05\% & -0.00 & 0.0409 \\
 &  & 3 & 6.33\% & 4.70\% & -0.00 & 0.0219 \\
 &  & 5 & 6.20\% & 4.94\% & -0.00 & 0.0224 \\ 
\cmidrule(lr){1-7}
\multirow{6}{*}{   Random  } & \multirow{3}{*}{2017-02} & 1 & 9.78\% & 7.63\% & 0.01 & 0.1099 \\
 &  & 3 & 10.02\% & 9.04\% & 0.01 & 0.1733 \\
 &  & 5 & 8.98\% & 8.78\% & 0.01 & 0.4032 \\ 
\cmidrule(lr){2-7}
 & \multirow{3}{*}{2018-02} & 1 & 9.88\% & 7.63\% & 0.01 & 0.0929 \\
 &  & 3 & 8.87\% & 8.17\% & 0.01 & 0.2420 \\
 &  & 5 & 8.40\% & 9.21\% & 0.02 & 0.8642 \\
\bottomrule
\end{tabular}
}
\end{table*}

\section{Ablation Study \& Analysis}
\subsection{How Many Samples Are Sufficient?}
\begin{figure}[t]
    \centering
    \includegraphics[width=1\linewidth]{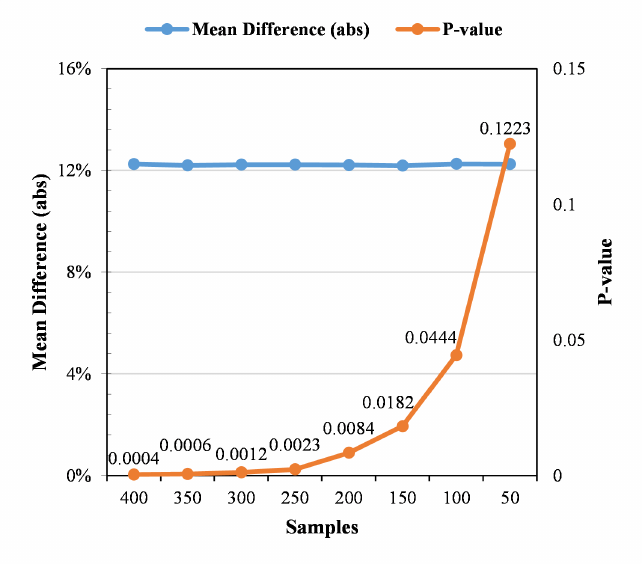}
    \caption{Mean difference and $p$-value on MC-PDD using LLaMA-3.1-8B-Instruct as functions of decreasing sample size. The $p$-value remains below 0.05 when the sample size exceeds 100.}
    \label{fig:sample}
\end{figure}

We investigated the minimum sample size required to maintain statistical significance in detection on LLaMA-3.1-8B-Instruct. Starting with 400 samples, we gradually reduced the sample size in increments of 50 and used bootstrap resampling \cite{efron1992bootstrap} to assess the significance of the hit rate difference (see Figure \ref{fig:sample}). Notably, the hit rate difference remains statistically significant even at a sample size of 100. 


We then reduce the sample size while monitoring the $p$-value and the upper bounds of the confidence intervals on the five closed-source models. The results in Table \ref{tab:main result samples} closely align with the trends observed in Figure \ref{fig:sample}. As expected, both the $p$-value and the upper bounds of the confidence intervals increase as the sample size decreases. Most models continue to exhibit a significant hit rate difference even at a sample size of 100, further supporting the effectiveness of our method. Notably, GPT-4o-Mini fails to maintain a significant difference by a small margin when the sample size drops to 150. Due to the model’s closed-source nature, a precise explanation for this behavior cannot be provided at this time.

\begin{table*}[t]
\centering
\caption{$P_{value}$ and upper bound of the confidence interval for five LLMs with decreasing sample size. Data highlighted in red indicate non-significant hit rate differences. An asterisk (*) denotes a knowledge cutoff later than December 2023. For models with a knowledge cutoff earlier than December 2023, the hit rate difference becomes non-significant when the sample size is reduced to 50, except for GPT-4o-Mini, where it becomes non-significant when the sample size falls below 150.}
\label{tab:main result samples}
\setlength{\extrarowheight}{0pt}
\addtolength{\extrarowheight}{\aboverulesep}
\addtolength{\extrarowheight}{\belowrulesep}
\setlength{\aboverulesep}{0pt}
\setlength{\belowrulesep}{0pt}
\resizebox{.98\textwidth}{!}{
\begin{tabular}{llcccccccc} 
\toprule
\multirow{2}{*}{LLM}                & \multirow{2}{*}{Bootstrap Result} & \multicolumn{8}{c}{Sample Size}                                                                                                                                                                                                                                                                                                                                                \\ 
\cmidrule(l){3-10}
& & 400                                         & 350                                            & 300                                            & 250                                         & 200                                       & 150                                            & 100                                         & 50                                           \\ 
\midrule
\multirow{2}{*}{Claude-3-Haiku}   & $P_{value}$        & 0.00006                                     & 0.00016                                        & 0.0003                                         & 0.00112                                     & 0.00314                                   & 0.00888                                        & 0.02792                                     & {\cellcolor[rgb]{0.827,0.725,0.718}}0.09776  \\
                                  & CI Upper Bound & -0.0825                                     & -0.0771                                    & -0.0733                                   & -0.068                                      & -0.06                                     & -0.0466                                   & -0.02                                       & {\cellcolor[rgb]{0.827,0.725,0.718}}0.02     \\
\multirow{2}{*}{Deepseek-V3-Chat} & $P_{value}$        & 0.00026                                     & 0.00074                                        & 0.00172                                        & 0.00392                                     & 0.00828                                   & 0.01954                                        & 0.04798                                     & {\cellcolor[rgb]{0.827,0.725,0.718}}0.12998  \\
                                  & CI Upper Bound & -0.0625                                     & -0.06                                          & -0.0533                                    & -0.048                                      & -0.04                                     & -0.0266                                   & -0.01                                       & {\cellcolor[rgb]{0.827,0.725,0.718}}0.04     \\
\multirow{2}{*}{GPT-4o-Mini}      & $P_{value}$        & 0.00502                                     & 0.00774                                        & 0.0135                                         & 0.02088                                     & 0.03554                                   & {\cellcolor[rgb]{0.827,0.725,0.718}}0.0622     & {\cellcolor[rgb]{0.827,0.725,0.718}}0.10874 & {\cellcolor[rgb]{0.827,0.725,0.718}}0.20576  \\
                                  & CI Upper Bound & -0.035                                      & -0.0286                                    & -0.0233                                   & -0.02                                       & -0.01                                     & {\cellcolor[rgb]{0.827,0.725,0.718}}0          & {\cellcolor[rgb]{0.827,0.725,0.718}}0.02    & {\cellcolor[rgb]{0.827,0.725,0.718}}0.08     \\
\multirow{2}{*}{LLaMA-3-70B}      & $P_{value}$        & 0.00006                                     & 0.00012                                        & 0.00024                                        & 0.0009                                      & 0.0029                                    & 0.00796                                        & 0.02668                                     & {\cellcolor[rgb]{0.827,0.725,0.718}}0.09344  \\
                                  & CI Upper Bound & -0.0825                                     & -0.08                                          & -0.0733                                    & -0.068                                      & -0.06                                     & -0.0467                                    & -0.03                                       & {\cellcolor[rgb]{0.827,0.725,0.718}}0.02     \\
\multirow{2}{*}{Gemini-1.5-Flash\textsuperscript{*}} & $P_{value}$        & {\cellcolor[rgb]{0.827,0.725,0.718}}0.30466 & {\cellcolor[rgb]{0.827,0.725,0.718}}0.31838    & {\cellcolor[rgb]{0.827,0.725,0.718}}0.33274    & {\cellcolor[rgb]{0.827,0.725,0.718}}0.35146 & {\cellcolor[rgb]{0.827,0.725,0.718}}0.37  & {\cellcolor[rgb]{0.827,0.725,0.718}}0.38984    & {\cellcolor[rgb]{0.827,0.725,0.718}}0.41948 & {\cellcolor[rgb]{0.827,0.725,0.718}}0.46362  \\
                                  & CI Upper Bound & {\cellcolor[rgb]{0.827,0.725,0.718}}0.0375  & {\cellcolor[rgb]{0.827,0.725,0.718}}0.0429 & {\cellcolor[rgb]{0.827,0.725,0.718}}0.0467 & {\cellcolor[rgb]{0.827,0.725,0.718}}0.056   & {\cellcolor[rgb]{0.827,0.725,0.718}}0.065 & {\cellcolor[rgb]{0.827,0.725,0.718}}0.0733 & {\cellcolor[rgb]{0.827,0.725,0.718}}0.1     & {\cellcolor[rgb]{0.827,0.725,0.718}}0.14     \\
\bottomrule
\end{tabular}}
\end{table*}

\subsection{How Much Information is Sufficient?}

In this step, we investigate the impact of removing the entity name from the prompt, providing only the entity’s information. Accordingly, the prompt template is modified as Figure \ref{fig:prompt_template}.

\begin{figure}[ht]
\centering
\small
\begin{tabular}{|p{0.45\textwidth}|}
\hline
\textbf{Prompt template} \\
\hline
Here is the introduction of an entity:\\
Introduction: \{introduction\}\\
List the 10 most likely candidate words for the \texttt{<blank>}, without any explanation:\\
\hline
\end{tabular}
\caption{Prompt template used for masked word prediction with entity information. Curly brackets \{\} denote placeholders to be completed.}
\label{fig:prompt_template}
\end{figure}


The test results, shown in Table \ref{tab: imformation}, reveal that our method still maintains a significant hit rate difference even when only entity information is provided. 
This demonstrates the method's applicability to datasets lacking explicit sample names. However, it is important to note that reducing the specificity of the input by removing the entity name substantially lowers the model's overall prediction hit rate for both member and non-member samples. In particular, the hit rate for member samples drops significantly by 15.59\%, indicating that the absence of entity names severely impairs the model’s memory activation.
These findings further confirm that specific information plays a crucial role in triggering the model’s memory of particular details, and that its removal causes the MC-PDD to function more as a test of general language modeling ability rather than memory recall.

\begin{table}[ht]
\centering
\caption{Hit rate and confidence interval upper bound on LLaMA-3.1-8B-Instruct. As the specificity of the information decreases, the hit rate on both member and non-member samples in LLaMA-3.1-8B-Instruct decreases. Particularly, the hit rate on member samples drastically decreased by 15.59\%.}
\begin{tabular}{lcc} 
\toprule
 & \multicolumn{1}{l}{Name \& Introduction} & \multicolumn{1}{l}{Introduction Only} \\
 \hline
Hit Rate on Non-members & 36.26\% & 26.07\% \\
Hit Rate on Members & 48.50\% & 32.91\% \\
Hit Rate Difference & -12.24\% ($P$=0.0001) & -6.84\% ($P$=0.0116) \\
CI Upper Bound & -7\% & -2\% \\
\bottomrule
\end{tabular}

\label{tab: imformation}
\end{table}

\subsection{The Impact of Number of Mask}
In this step, we investigate whether selecting low-specificity words affects the results. To this end, we expand the TF-IDF selection size, including less specific words from the same samples for masking and prompting the LLM to predict them. The results, shown in Table \ref{tab:num of mask} 
, indicate that incorporating less specific words into the queries reduces the hit rate difference and decreases statistical significance. This supports our hypothesis that low-specificity words reduce the hit rate difference, turning the MC-PDD into more of a language modeling test rather than a memorization assessment.

Despite these promising results, we remain cautious about the effectiveness of TF-IDF for selecting specific words across different contexts. While simple and efficient, TF-IDF relies solely on word frequency and does not account for semantics or part-of-speech, which may lead to the omission of truly distinctive words. We plan to explore alternative selection methods in future experiments.  

\begin{table*}[ht]
\centering
\caption{Results of the MC-PDD after expand number of mask. The hit rate on both member and non-member samples decreases as the number of masked tokens increases, and the hit rate difference between them continues to diminish.}
\label{tab:num of mask}
\resizebox{.9\textwidth}{!}{
\begin{tabular}{ccccc} 
\toprule
\multicolumn{1}{l}{Mask} & \multicolumn{1}{l}{Hit Rate on Non-members} & \multicolumn{1}{l}{Hit Rate on Members} & \multicolumn{1}{l}{Hit Rate Difference} & \multicolumn{1}{l}{$P-value$}  \\
\cmidrule(){1-5}
1                        & 36.26\%                                     & 48.50\%                                 & -12.24\%                                & 0.00016                      \\
2                        & 34.24\%                                     & 42.09\%                                 & -7.85\%                                 & 0.00048                      \\
3                        & 32.78\%                                     & 38.96\%                                 & -6.18\%                                 & 0.00104                      \\
\bottomrule
\end{tabular}}
\end{table*}

\subsection{Impact of Data Contamination}
\begin{figure}[!ht]
    \centering
    \includegraphics[width=.98\linewidth]{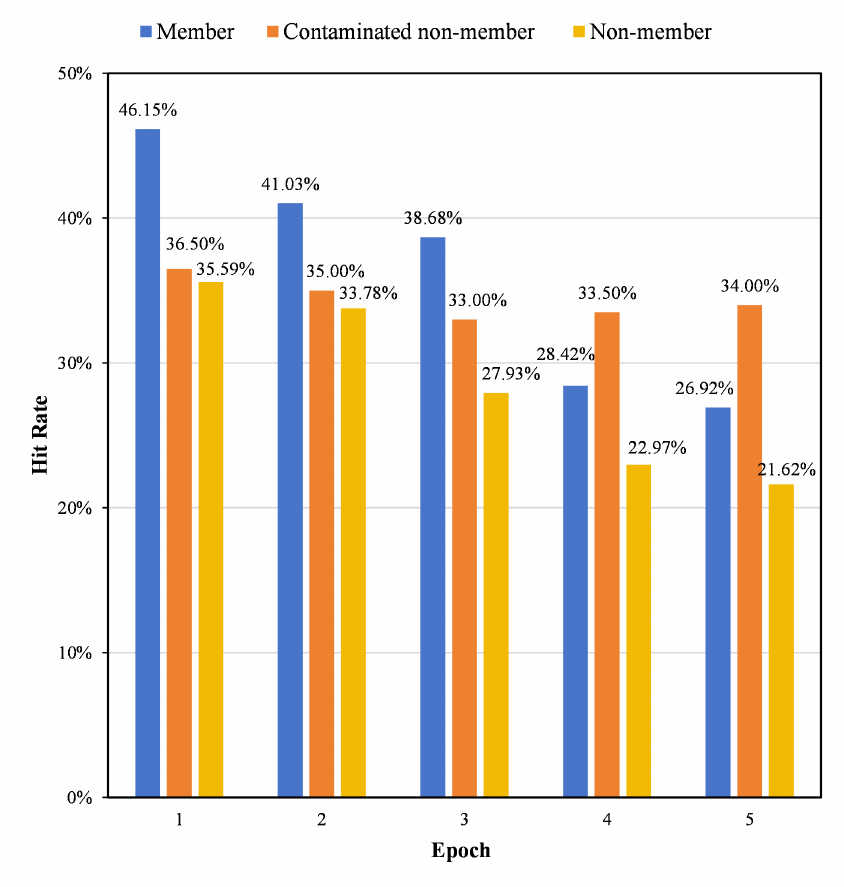}
    \caption{Hit rates for member, non-member, and contaminated non-member samples across training epochs.}
    \label{fig:epoch}
\end{figure}

Finally, to simulate potential contamination in pretraining, we randomly select 200 non-member samples from SteamMIA for additional training of LLaMA-3.1-8B-Instruct and performed MC-PDD after each epoch. As shown in Figure \ref{fig:epoch}, the results reveal a steady decline in the model’s hit rate for both member and non-member samples as training progresses. We attribute this trend to the negative effects of continued training on a model that has already undergone instruction fine-tuning, which can degrade both its instruction-following ability and language modeling capability. 
Interestingly, the model's hit rate on contaminated non-member samples initially decreases slightly, followed by a modest increase. This pattern suggests that the model’s memory of the contaminated non-member samples can be reinforced through continued training, indirectly validating the effectiveness of our method.

\section{Conclusion}
This paper introduces MC-PDD, a masked corpus-level black-box method for pretraining data detection. Unlike existing approaches that operate at the sample level, MC-PDD elevates detection to the corpus level and requires only model outputs, significantly enhancing its practicality. We also experimentally determine the minimum sample size necessary for reliable corpus-level detection.
The core idea of MC-PDD is to activate the model’s memory of highly specific information that is difficult to infer. By guiding the model to predict masked words selected based on TF-IDF scores, we mitigate the impact of the model’s generalization capability. This approach provides a simple yet effective solution for corpus-level pretraining data detection.
To support experimental validation, we constructed SteamMIA, a benchmark comprising publicly available game data from Steam. Furthermore, we evaluated MC-PDD on BBC\_News\_Alltime and ArXiv\_Alltime, demonstrating its robust generalizability across different data domains.

We evaluated our method through extensive experiments under multiple settings. First, we tested MC-PDD on both open-source and closed-source models, benchmarking its effectiveness against prior sample-level detection methods. Second, we examined the minimum sample size required for stable detection. Third, we investigated the effects of data release date, word selection strategy and selection size, and dataset contamination on detection performance. Collectively, these experiments demonstrate the effectiveness and robustness of MC-PDD from multiple perspectives.
We will release SteamMIA to facilitate reproducibility and support future MIA research, and plan to optimize specificity word selection and extend MC-PDD to document-level detection.

\section*{Limitations}

Although MC-PDD demonstrates strong efficiency, practicality, and robustness, several limitations remain:

(i) \textbf{Scalability and Model Diversity}: Due to time and computational constraints, MC-PDD has not been tested on larger-scale models or additional public datasets. Further validation is needed to assess its effectiveness on more diverse model architectures and training approaches.

(ii) \textbf{Specificity Requirements}: MC-PDD relies on corpora containing a sufficient number of specificity words. When applied to corpora dominated by common everyday language, the effectiveness of specificity-based word selection may be diluted, potentially reducing detection accuracy.

(iii) \textbf{Word Selection Strategy}: The current approach selects specificity words primarily based on frequency, which may overlook key terms carrying meaningful information. Incorporating richer linguistic features, such as semantic information or part-of-speech analysis, may further enhance performance.


\section*{Acknowledgements}
This work was supported in part by the Science and Technology Development Fund of Macau SAR (Grant Nos. FDCT/0007/2024/AKP, EF2024-00185-FST), the UM and UMDF (Grant Nos. MYRG-GRG2024-00165-FST-UMDF, MYRG-GRG2025-00236-FST), the Tencent AI Lab Rhino-Bird Research Program (Grant No. EF2023-00151-FST), the Dr. Stanley Ho Medical Development Foundation (Grant No. SHMDF-AI/2026/001), and the National Natural Science Foundation of China (Grant No. 62266013). This work was performed in part at SICC which is supported by SKL-IOTSC, and HPCC supported by ICTO of the University of Macau.

\bibliographystyle{IEEEtran}
\bibliography{custom}

\end{document}